\pdfoutput=1

\documentclass[11pt]{article}

\usepackage{acl}

\usepackage{times}
\usepackage{latexsym}
\usepackage{amsmath}
\usepackage{amssymb}
\usepackage{bbm}
\usepackage{dsfont}

\usepackage[T1]{fontenc}

\usepackage[utf8]{inputenc}

\usepackage{microtype}
\usepackage{times}
\usepackage{latexsym}
\usepackage{soul}
\usepackage{amsmath}
\usepackage{booktabs}
\usepackage{bm}
\usepackage{cleveref}
\usepackage{graphicx}
\usepackage{tabularx}
\usepackage{soul}
\usepackage{xcolor}
\usepackage{todonotes}
\usepackage{multirow}
\usepackage{xpatch}
\usepackage{blindtext}
\usepackage{fdsymbol}
\usepackage{microtype}
\usepackage{hyperref}
\usepackage{adjustbox}
\usepackage{textcomp}
\usepackage{xparse}
\usepackage{enumitem}
\usepackage{makecell}
\usepackage{subcaption}
\usepackage{colortbl}
\usepackage{array}

\usepackage{xparse}
\usepackage{highlight}
\usepackage{todonotes}
\usepackage{stfloats}

\crefformat{section}{\S#2#1#3} 
\crefformat{subsection}{\S#2#1#3}
\crefformat{subsubsection}{\S#2#1#3}

\title{\textsc{AMRFact}: Enhancing Summarization Factuality Evaluation\\with AMR-Driven Negative Samples Generation}

\author{Haoyi Qiu\textsuperscript{$\heartsuit$} ~~ Kung-Hsiang Huang\textsuperscript{$\diamondsuit$}$^*$ ~~ Jingnong Qu\textsuperscript{$\heartsuit$}$^*$ ~~ Nanyun Peng\textsuperscript{$\heartsuit$} \\
\textsuperscript{$\heartsuit$}University of California, Los Angeles ~~~
\textsuperscript{$\diamondsuit$}University of Illinois Urbana-Champaign \\
\textsuperscript{$\heartsuit$}\texttt{\{haoyiqiu, jingnong.qu, violetpeng\}@cs.ucla.edu} \\
\textsuperscript{$\diamondsuit$}\texttt{khhuang3@illinois.edu} \\}

\begin{document}
\maketitle
{\def\thefootnote{*}\footnotetext{The authors contributed equally to this work and are listed in alphabetical order by last name.}}
\newcommand{\samsum}[1]{\textsc{SAMSum}}
\newcommand{\dialogsum}[1]{\textsc{DialogSum}}
\newcommand{\mixandmatch}[1]{\textsc{MixAndMatch}}
\newcommand{\confit}[1]{\textsc{ConFiT}}
\newcommand{\ctrldiasumm}[1]{\textsc{CtrlDiaSumm}}
\newcommand{\cods}[1]{\textsc{CODS}}
\newcommand{\modelshort}[1]{\textsc{AMRFact}}
\newcommand{\aggrefactsota}[1]{\textsc{AggreFact-FtSota}}
\newcommand{\cnndm}[1]{\textsc{CNN/DM}}
\newcommand{\xsum}[1]{\textsc{XSum}}
\newcommand{\negfilter}[1]{\textsc{NegFilter}}

\newcommand{\datashort}[1]{\textsc{DiverseSumm}}
\newcommand{\tasklong}[1]{Multi-document Diversity Summarization}
\newcommand{\taskshort}[1]{MDDS}
\newcommand{\gptturbo}[1]{\texttt{gpt-3.5-turbo}}
\newcommand{\gptturbolong}[1]{\texttt{gpt-3.5-turbo-16k}}
\newcommand{\gptfour}[1]{\texttt{gpt-4}}
\newcommand{\vicuna}[1]{\texttt{vicuna-7b}}
\newcommand{\longchat}[1]{\texttt{longchat-7b-16k}}
\newcommand{\CC}[1]{\cellcolor{lightblue!#1}}

\definecolor{c2}{RGB}{218,0,0}
\newcommand{\propaHighlight}[1]{{\color{c2} {#1}}}
\definecolor{lightblue}{RGB}{212, 235, 255}
\definecolor{lightorange}{RGB}{255, 204, 168}
\definecolor{lightyellow}{RGB}{255, 255, 168}
\definecolor{lightred}{RGB}{255, 168, 168}
\definecolor{lightpink}{RGB}{248, 206, 204}
\definecolor{darkred}{RGB}{196, 30, 58}
\definecolor{lightgreen}{RGB}{213, 232, 212}
\definecolor{lightgreen}{rgb}{0.82, 0.94, 0.75}
\definecolor{lightpurple}{RGB}{225, 213, 231}
\definecolor{darkgreen}{rgb}{0.56, 0.63, 0.51}
\definecolor{normalgreen}{rgb}{0.66, 0.9, 0.5}
\definecolor{lightgray}{rgb}{0.7, 0.7, 0.7}
\definecolor{gold}{rgb}{0.83, 0.69, 0.22}
\sethlcolor{lightblue}
\newcolumntype{Y}{>{\centering\arraybackslash}X}
\newcommand\hlc[2]{\sethlcolor{#1} \hl{#2}}
\iftrue

\NewDocumentCommand{\steeve}
{ mO{} }{\textcolor{gold}{\textsuperscript{\textit{Steeve}}\textsf{\textbf{\small[#1]}}}}

\else
\newcommand{\steeve}[1]{}

\fi
\definecolor{gold}{rgb}{0.83, 0.69, 0.22}

\newcommand{\Steeve}[1]{{\color{orange}#1}}
\newcommand{\markred}[1]{{\color{darkred}#1}}
\newcommand{\cmark}{\ding{51}}%
\newcommand{\xmark}{\ding{55}}%

\newcommand{\haoyi}[1]{\textcolor{blue}{Haoyi: #1}}

\newcommand{\SideNote}[2]{\todo[color=#1,size=\small]{#2}} 
\newcommand{\violet}[1]{\SideNote{purple!40}{#1 --violet}}
\begin{abstract}
Ensuring factual consistency is crucial for natural language generation tasks, particularly in abstractive summarization, where preserving the integrity of information is paramount. Prior works on evaluating factual consistency of summarization often take the entailment-based approaches that first generate perturbed (factual inconsistent) summaries and then train a classifier on the generated data to detect the factually inconsistencies during testing time. However, previous approaches generating perturbed summaries are either of \textit{low coherence} or \textit{lack error-type coverage}. To address these issues, we propose \modelshort~, a framework that generates perturbed summaries using Abstract Meaning Representations (AMRs). Our approach parses factually consistent summaries into AMR graphs and injects controlled factual inconsistencies to create negative examples, allowing for coherent factually inconsistent summaries to be generated with high error-type coverage. Additionally, we present a data selection module \negfilter~ based on natural language inference and \textsc{BARTScore} to ensure the quality of the generated negative samples. Experimental results demonstrate our approach significantly outperforms previous systems on the \aggrefactsota~ benchmark, showcasing its efficacy in evaluating factuality of abstractive summarization.\footnote{Data and code are available at \url{https://github.com/PlusLabNLP/AMRFact}}

\end{abstract}
\section{Introduction}
\vspace{-2mm}

\begin{figure}[t]
 \centering
 \includegraphics[width=\linewidth]{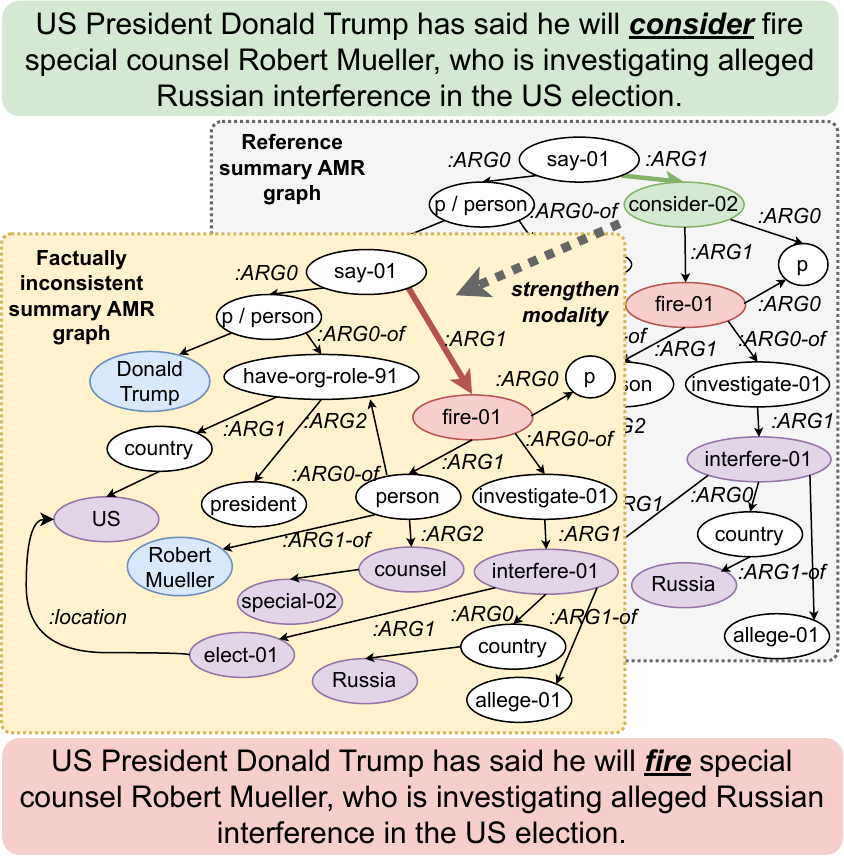}
 \vspace{-7mm}
 \caption{Example of a reference (green) and a generated factually inconsistent summary (red) from the \modelshort~ dataset. Given a reference summary, we convert the text into an AMR graph (grey) and then remove ``\hlc{lightgreen}{consider-02}'' to generate a factually inconsistent summary AMR graph (yellow). This perturbed summary strengthens the modality in the reference summary, resulting in factual inconsistency. The reference and perturbed summaries will be used as positive and negative examples, respectively.}
 \label{fig:amrfact_intro}
 \vspace{-7mm}
\end{figure}

\begin{figure*}[t]
 \centering
 \includegraphics[width=\textwidth]{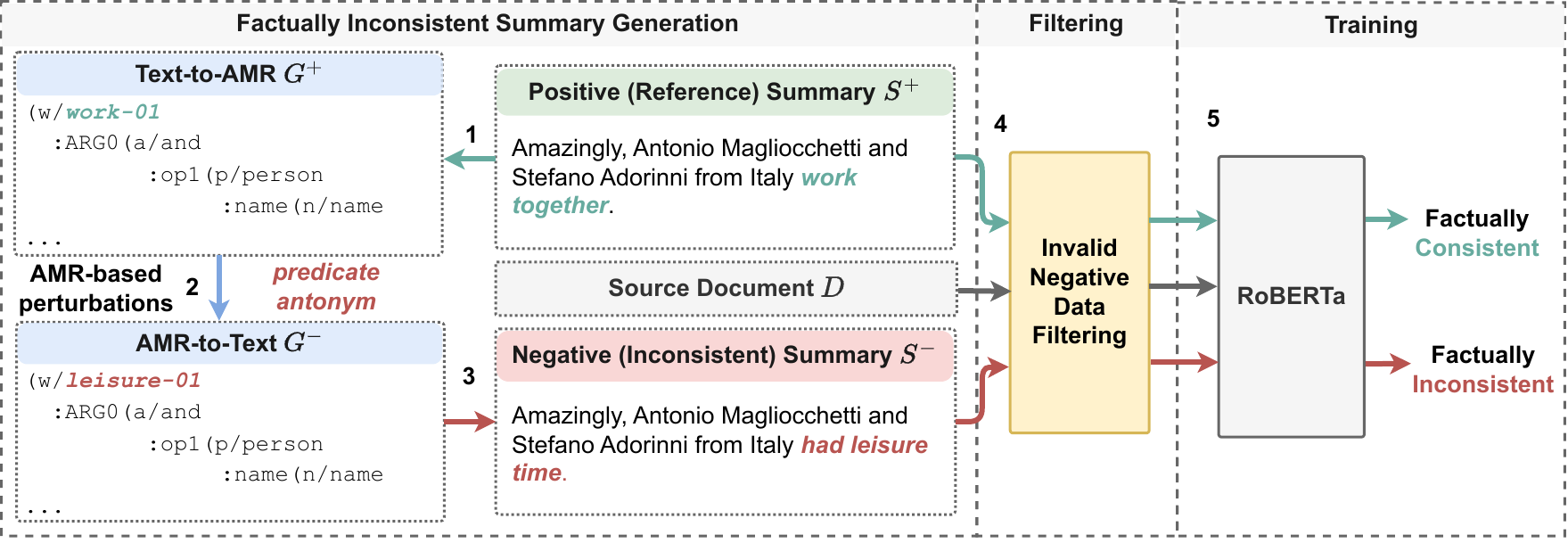}
 \vspace{-7mm}
 \caption{Overview of \textsc{AMRFact} training phase: (1) The generation module first converts the reference summaries into AMR graphs. (2) These graphs are then manipulated to include common factual errors shown in current summarization systems, creating factually inconsistent AMR graphs. (3) These manipulated graphs are back-translated into text summaries, serving as negative examples for training a text-based factuality evaluator. (4) A selection module, using NLI score and \textsc{BARTScore}, filters out low-quality negative examples. (5) Finally, we fine-tune a \textsc{RoBERTa}-based model with this data to act as the \textit{evaluation metric}, assessing factuality by comparing the original document (premise) with the summary (hypothesis) and measuring the probability of entailment.}
 \vspace{-5mm}
 \label{fig:amrfact_overview}
\end{figure*}

Recent advances in text summarization, driven by pre-trained language models, have enabled the generation of coherent abstractive summaries \cite{Zhang2019PEGASUSPW, Raffel2019ExploringTL, lewis-etal-2020-bart}. However, studies have shown that these generated summaries can be factually inconsistent with the source documents \cite{Goodrich2019AssessingTF,kryscinski-etal-2020-evaluating,pagnoni-etal-2021-understanding, huang-etal-2023-do, kim-etal-2023-can}. This inconsistency between the generated summary and the factual information in the source document necessitates the need for assessing the \textit{factuality} or factual consistency of the summaries.

Recent work formulated factual inconsistency detection as an entailment recognition task, predicting whether the source document entails a summary. The prevalent approach for developing an entailment-based factual consistency metric involves generating synthetic data and training a classifier on it. \citet{kryscinski-etal-2020-evaluating} treats reference summaries from \cnndm~ \cite{Hermann2015TeachingMT} as positive examples and applies entity-centric perturbations, such as entity replacement, to the reference summaries to create negative samples (\textit{i.e.}, summaries factually inconsistent with the source document). However, such an approach often produces sentences with poor coherence caused by the string replacement operations. \citet{goyal-durrett-2021-annotating} and \citet{utama-etal-2022-falsesum} address the coherence issue by training a paraphrasing model and a text-infilling model to produce negative data. Yet, these approaches cannot ensure the production of specific types of factual errors illustrated in \citet{pagnoni-etal-2021-understanding}. As a result, models trained on data produced by these methods may not reliably detect particular types of factual inconsistency in summaries. We refer to this limitation as a lack of ``\textit{error-type coverage}''.
Besides, all prior methods lack a verification step to \textit{validate} the quality of the generated data, potentially leading to diminished performance in detecting factual inconsistency. 

Motivated by these challenges, we propose \modelshort~, a framework that generates \textbf{coherent negative examples} with \textbf{high error-type coverage}. Our data generation module leverages Abstract Meaning Representations (AMRs) \cite{Banarescu2013AbstractMR} to introduce semantic-level perturbations for creating negative examples, enabling us to generate more coherent summaries without compromising the error-type coverage. AMRs are intended to capture the meaning of a sentence by abstracting away irrelevant syntactic features. This feature allows us to maintain coherence while precisely tailoring our negative examples to target specific factual error types. As shown in prior research \cite{Lee2021AnAO}, AMR's \textit{controllability} empowers us to easily shape the distribution of negatives, making our approach highly adaptable and effective in addressing the identified challenges.

In detail, \modelshort~ starts with parsing factually consistent summaries into semantic AMR graphs, and then injects factual inconsistencies (errors) that are \textit{commonly} observed in state-of-the-art summarization systems \cite{pagnoni-etal-2021-understanding} into the AMR graphs \cite{ghazarian-etal-2022-deam, ribeiro-etal-2022-factgraph}. These perturbed AMR graphs are subsequently translated back into text summaries to serve as our negative examples using a controllable generation model, which ensures that the generated summaries retain a \textit{natural} and \textit{coherent} narrative flow. Figure~\ref{fig:amrfact_intro} shows an example of such AMR-based data generation. 
Then, we devise a novel selection module \negfilter~ to exclude invalid negative samples from our training data. A \textit{valid} negative summary must satisfy two criteria: (1) it must not be directly inferable from the original summary and (2) it should not stray significantly from the main topic of the document. We employ sentence-level NLI and \textsc{BARTScore} to ensure compliance with each criterion, respectively. Finally, a \textsc{RoBERTa} \cite{liu2019roberta} model is trained on the created dataset along to distinguish factually consistent and inconsistent summaries as the evaluation metric. \Cref{fig:amrfact_overview} offers an overview of our proposed framework. The highlights of our contributions include:\looseness=-1

\vspace{-1mm}

\begin{itemize}
    \item We propose \modelshort~ that uses AMR-based perturbations to generate factually inconsistent summaries, which allows for more coherent generation with high error-type coverage.
    \item We devise a data validation module \negfilter~ to filter out invalid negative summaries to enhance the quality of generated data.
    \item Our approach achieves state-of-the-art performance on the \aggrefactsota~ dataset.
\end{itemize}

\begin{figure*}[t]
 \centering
 \includegraphics[width=\textwidth]{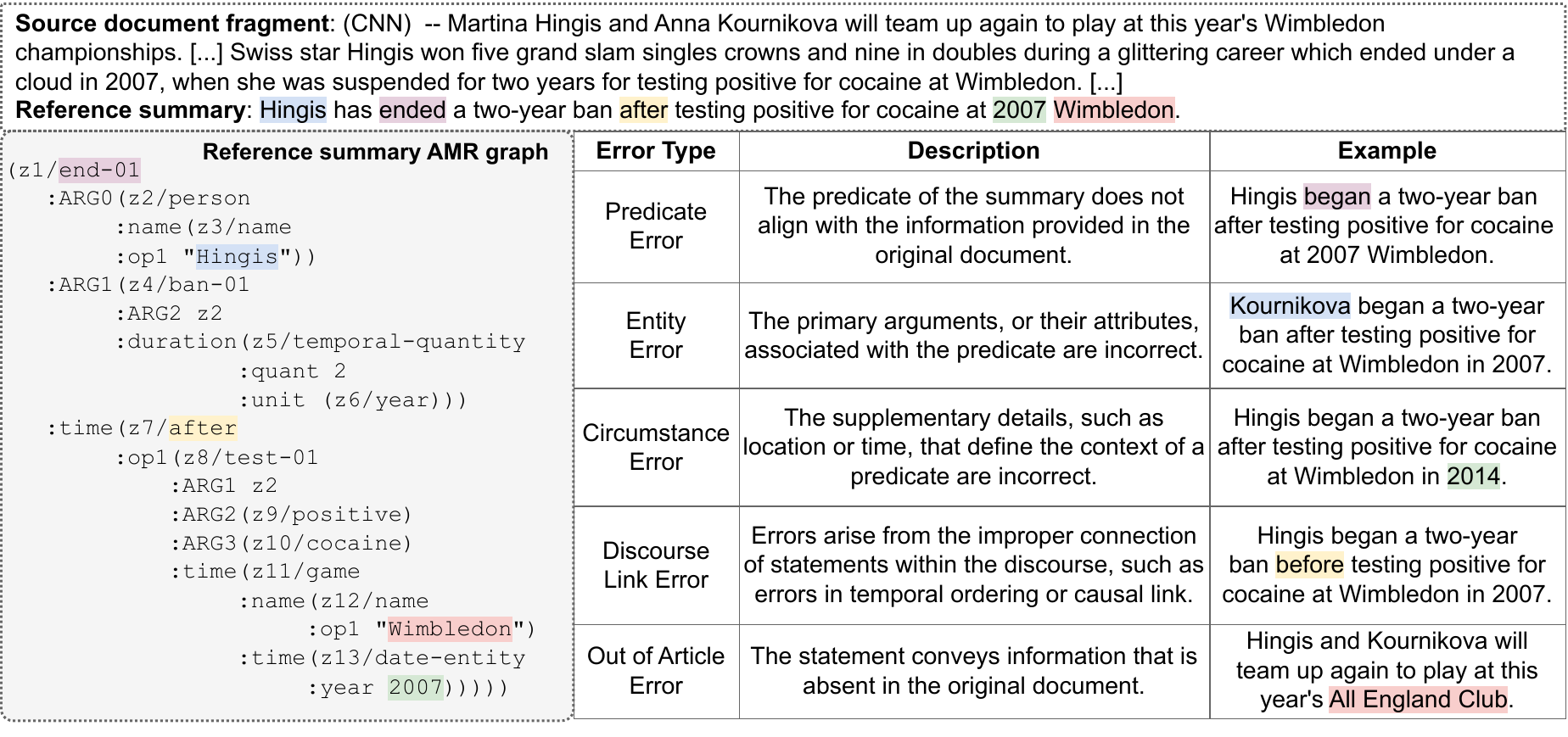}
 \vspace{-9mm}
 \caption{Typology of factual errors. Given the source document and reference summary, we apply five kinds of factual inconsistencies: predicate error, entity error, circumstance error, discourse link error, and out-of-article error. Each color represents the implementation of one kind of factual error from reference summary to perturbed summary.\looseness=-1}
 \label{fig:amrfact_errors_typology}
 \vspace{-5mm}
\end{figure*}

\vspace{-5mm}
\section{Abstract Meaning Representations}
\vspace{-1mm}

Introduced by \citet{Banarescu2013AbstractMR}, AMR is a representation language that effectively captures the essence of texts by encoding abstract-level semantic information, including named entities, negations, coreferences, and modalities. This intrinsic capability positions AMR as a critical asset for various semantic-related NLP tasks, such as summarization \cite{liao2018abstract} and machine translation \cite{song-etal-2019-semantic}. In our study, we harness the potential of AMR in the summarization factuality evaluation task by perturbing the graphs of factually consistent summaries. \textbf{Each perturbation reflects a factual error in summarization.} In AMR graphs, nodes symbolize entities and concepts, connected through different relational edges. Figure~\ref{fig:amrfact_intro} provides an example of an AMR graph for a summary.\looseness=-1

\section{\modelshort~}

In \modelshort~, we start by parsing reference summaries into AMR graphs and then introduce factual errors into these graphs. Subsequently, we back-translate the perturbed AMR graphs into text and use data filtering to acquire high-quality negative training samples. Finally, we combine the positive samples (reference summaries) with the negatives to train a classifier, which serves as our factuality evaluation metric. More specifically, the process of \textit{generating} factually inconsistent summaries involves two primary states: introducing summary-level perturbations based on AMR and invalid negative data filtering. They are discussed in more detail in the following sections dedicated to explaining how factual errors are introduces (\Cref{sec:factual-errors}) and the criteria for choosing valid negative examples (\Cref{sec:negative-examples-selection}). \looseness=-1

\vspace{-2mm}
\subsection{AMR-based Summary Perturbations}
\label{sec:factual-errors}

Our approach targets five types of \textit{factual inconsistencies}: predicate error, entity error, circumstance error, discourse link error, and out-of-article error, as categorized by \citet{pagnoni-etal-2021-understanding}\footnote{We do not discuss or implement coreference errors because this work targets sentence-level summaries, and it is hard for them to contain pronouns or references with wrong or nonexisting antecedents.}. A detailed description of these factual errors is presented in \Cref{fig:amrfact_errors_typology}. Each type of errors can be generated by perturbing the AMR graph of a summary that is initially factually consistent, and then reconverting the perturbed graph back into a natural language summary. Specifically, we start with a source document $D$ alongside its factually consistent summary $S^+$, which is then parsed into an acyclic AMR graph $G^+$. This graph is manipulated to form a new AMR graph $G^-$ that embodies factual inconsistencies. For instance, as illustrated in \Cref{fig:amrfact_overview}, we produce a predicate error by swapping the original predicate ``work'' to ``leisure''. The perturbed AMR graph $G^-$ is then transformed into a factually inconsistent natural language summary $S^-$ using an AMR-to-Text model \cite{ribeiro-etal-2021-investigating}\footnote{We utilize the Text-to-AMR and AMR-to-Text models from https://github.com/bjascob/amrlib-models.}.

\paragraph{Predicate Error.} Predicate errors occur when the predicate in a summary does not align with the information in the source document. We simulate this type of error based on two processes: (1) by adding or removing polarity and (2) through the substitution of a predicate with its antonym. By directly adding or removing \textit{polarity} to the concpets, we change the negation in the sentence. Another approach is the antonym substitution. Here, we replace the concepts with their antonyms that holds \textit{Antonym}, \textit{NotDesires}, \textit{NotCapableOf}, and \textit{NotHasProperty} relations in ConceptNet \cite{speer-havasi-2012-representing}, and therefore modify the sentence-level relations.

\paragraph{Entity Error.} Entity errors manifest when the entities associated with a predicate in a summary are incorrectly attributed or erroneous. These errors are crafted through two principal sources: (1) by swapping the roles of the agent and the patient, which results in the misattribution of actions or characteristics, and (2) by substituting specific entities, such as names and numbers. In AMR graphs, the clear distinction between agent (\textit{ARG0}) and patient (\textit{ARG1}) allows for straightforward swaps. We implement agent-patient swaps by exchanging the roles of the agent and the patient. Here, the agent refers to an action doer, and the patient refers to an action recipient. We randomly choose a verb from the sentence in our implementations. For named entity modifications, we identify the nodes holding roles such as \textit{name} and \textit{quant} and then randomly select one to be replaced by a different node with the same role from the same source document. 

\paragraph{Circumstance Error.} Circumstance errors in summaries emerge when there is incorrect or misleading information regarding the context of predicate interactions, specifically in terms of location, time, and modality. These errors are mainly created in two ways: (1) by intensifying the modality, which alters the degree of certainty or possibility expressed in the statement, and (2) by substituting specific entities like locations and times. We strengthen the modality by replacing the concepts that controls modals (\textit{e.g.}, \textit{permit} $\rightarrow$ \textit{obligate}). For entity substitution, we identify the nodes holding roles such as \textit{location}, \textit{date}, and \textit{time} and then randomly select one to be replaced by a different node with the same role from the same source document.

\paragraph{Discourse Link Error.} 
Discourse link errors pertain to mistakes in the logical connections between various statements in a summary. We focus on two fundamental types of discourse links: (1) temporal order, which deals with the sequence of events, and (2) causality, which pertains to the cause-and-effect relationships between statements. To manipulate temporal ordering, we perturb the nodes holding concepts such as \textit{before}, \textit{after}, or \textit{now}. For causality modifications, we alter argument structures associated with \textit{cause}, either at the root or as a modifier, effectively reversing the causal relationship.

\paragraph{Out of Article Error.} 
Summaries are expected to contain only information that can be inferred from the source document, and deviations from this rule need to be clearly identified. To create an ``out of article'' error, we follow a similar method as previously discussed, involving alterations in entities, times, or locations. However, in this instance, we intentionally introduce vocabulary \textit{not} present in the original document. 

\subsection{Invalid Negative Data Filtering}
\label{sec:negative-examples-selection}

In our data generation process, we noticed that certain generated negative examples either blatantly contradicted the source article's facts or failed to modify the semantics of the reference summary, inadvertently becoming \textit{positive} examples.

These were deemed as \textit{invalid} negative examples. We hypothesize that \textbf{training detection models on such invalid negative data could potentially impair their performance}. To address this issue, we introduce a module \negfilter~ specifically designed for filtering out invalid negative data.

\begin{figure}[t]
    \centering
    \includegraphics[width=0.9\linewidth]{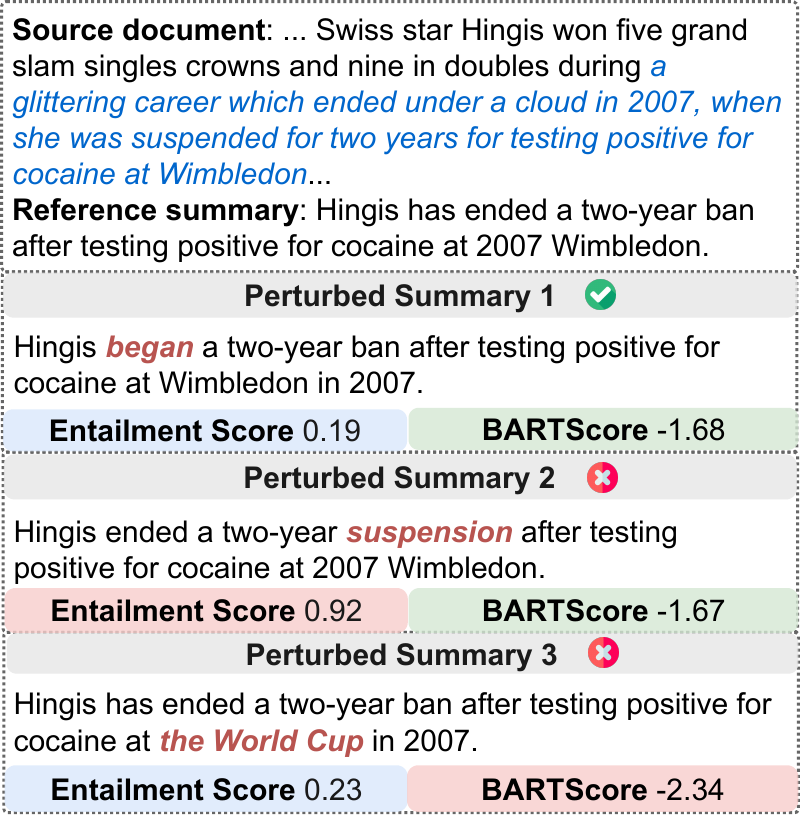}
    \caption{An illustration showing how our invalid negative data filtering module works. In the above three examples, only the first perturbed summary is valid since both of its entailment score and \textsc{BARTScore} satisfy the criteria described in \Cref{sec:negative-examples-selection}.} 
    \label{fig:amrfact_selection}
    \vspace{-4mm}
\end{figure}

A perturbed summary is considered \textit{valid} only if it satisfy two essential criteria: (1) it must be sufficiently distinct from the original summary to avoid being mistaken for a mere variation, thereby ensuring it is not misclassified as a positive example; and (2) despite the introduced perturbations, it should maintain a discernible connection to the source document, without diverging excessively.

To ensure the first criterion -- \textit{distinctiveness of generated summaries} -- we utilize sentence-level NLI inspired by previous studies \cite{wan-bansal-2022-factpegasus,huang-etal-2023-zero, huang-etal-2023-faking}. Concretely, we employ a RoBERTa large model fine-tuned on the MNLI corpus \cite{liu2019roberta, williams-etal-2018-broad}\footnote{\label{roberta-large} https://huggingface.co/roberta-large-mnli} to quantify the entailment score between an original summary ($S^+$) and its perturbed counterpart ($S^-$). Then, perturbed summaries that score exceed an empirically selected threshold ($\tau_1$) in entailment scores are discarded, given their elevated likelihood of being inferred from $S^+$. 

For the second criterion -- \textit{maintaining relevance to the source document} -- we propose to use \textsc{BARTScore} \cite{NEURIPS2021_e4d2b6e6}, fine-tuned on the \cnndm~ dataset \cite{Hermann2015TeachingMT}, to assess the semantic alignment between the perturbed summary ($S^-$) and its corresponding source document ($D$). Similarly, perturbed summaries with a \textsc{BARTScore} falling below the empirically determined threshold ($\tau_2$) are excluded due to their divergence from the source documents.

In summary, our proposed high-quality negative examples selection module \negfilter~ will take a source document ($D$), an original summary ($S^+$), and a perturbed summary ($S^-$). Only when both criteria are satisfied, the generated factually inconsistent summaries ($S^-$) will be included into our negative training examples. Formally,

\vspace{-4mm}
{\small 
\begin{gather}
\mathcal{M}(D, S^+, S^-) = \textbf{True} \text{ if and only if } \nonumber~\\
\mathbbm{1}[\mathcal{N}(S^+, S^-) < \tau_1] \cdot \mathbbm{1}[\mathcal{B}(D, S^-) > \tau_2] = 1,
\end{gather}
}
where $N$ represents the entailment score, and $B$ denotes \textsc{BARTScore}. The selection of thresholds is discussed in detail in \Cref{sec:threshold}. \Cref{fig:amrfact_selection} provides an illustration of our proposed selection module.

\vspace{-1mm}
\subsection{Detecting Factual Inconsistency}

To learn to detect factual inconsistency, we fine-tune a \texttt{roberta-large} model pre-trained on the MNLI corpus on our \textsc{AMRFact} dataset.\textsuperscript{\ref{roberta-large}} We directly adopt this pretrained model because the goal of natural language inference is very similar to that of factual inconsistency detection, \textit{i.e.}, predicting whether a given summary hypothesis ($S^+$ or $S^-$) can be entailed (factually consistent) with the given document premise ($D$). Although our output linear layer performs a three-way classification, since we only focus on two-way classification, \textit{i.e.}, factual consistency and inconsistency, we will only pay attention to the ``entailment'' and ``contradiction'' outputs. The training data paired with the reference summaries are labeled as ``entailment,'' whereas those coupled with the perturbed summaries are labeled as ``contradiction.'' \looseness=-1

\begin{table*}
\small
\centering
\begin{adjustbox}{max width=0.97\linewidth}
{
\begin{tabular}{lccccccccc}
\toprule
\multicolumn{2}{c}{} & Polytope  & SummEval & FRANK & Wang'20 & CLIFF & Goyal'21 & Cao'22 & Total \\
\midrule
\multirow{2}{*}{\textsc{AggreFact-Cnn-FtSota}} & val & 34 & 200 & 75 & - & 150 & - & - & 459 \\
                              & test & 34 & 200 & 175 & - & 150 & - & - & 559 \\
\midrule
\multirow{2}{*}{\textsc{AggreFact-Xsum-FtSota}} & val & - & - & - & 120 & 150 & 50 & 457 & 777 \\
                               & test & - & - & - & 119 & 150 & 50 & 239 & 558 \\
\bottomrule
\end{tabular}
}
\end{adjustbox}
\vspace{-2mm}
\caption{Statistics of \textsc{AggreFact-Cnn-FtSota} and \textsc{AggreFact-Xsum-FtSota}. These two subsets consist of Polytope \cite{huang-etal-2020-achieved}, SummEval \cite{fabbri2021summeval}, FRANK \cite{pagnoni-etal-2021-understanding}, Wang'20 \cite{wang-etal-2020-asking}, CLIFF \cite{cao-wang-2021-cliff}, Goyal'21 \cite{goyal-durrett-2021-annotating}, Cao'22 \cite{cao-etal-2022-hallucinated}. Each number in the table represents the number of samples.}
\vspace{-5mm}
\label{tab:aggrefact_stats}
\end{table*}

\section{Experimental Settings}

\subsection{Datasets}
\label{sec:dataset}

\paragraph{Training Dataset}
We apply the \textsc{AMRFact} data generation pipeline to the training split of the \cnndm~ corpus \cite{Hermann2015TeachingMT}, compromising English news articles and their associated human-generated summaries. In our approach to generating negative training data, we start with segmenting reference summaries into individual sentences. To finish this step efficiently, we reuse sentences from the positive samples of the \textsc{FalseSum} \cite{utama-etal-2022-falsesum} training dataset. This process yields 143,942 high-quality positive examples. Subsequently, we employ five distinct types of AMR-based perturbations on these positive examples. The perturbation process was exhaustive, meaning each positive example could lead to \textit{zero} to \textit{several} negative examples. After applying perturbations, we obtain 141,570 negative samples. These negative samples are then filtered by \negfilter~ and combined with the positive examples (reference summaries) to form our training dataset. Eventually, our dataset consists of 13,834 training and 2,000 validation instances, where we ensure an equal number of positive and negative data. Specifically, the dataset contains 35.85\% predicate errors, 33.75\% entity errors, 10.55\% circumstance errors, 3.7\% discourse link errors, and 16.16\% errors that are out of the article's scope.

\paragraph{Evaluation Dataset} 
\citet{tang-etal-2023-understanding} introduced the benchmark \textbf{\textsc{AggreFact}}, which consolidates \textit{nine} existing datasets focused on factuality for a finer-grained comparison of factuality assessment systems. They stratify the benchmark based on the underlying summarization model, categorized into \textsc{FtSota}, \textsc{EXformer}, and \textsc{Old} according to their development timeline. \textsc{FtSota} encompasses state-of-the-art fine-tuned summarization models, such as BART \cite{lewis-etal-2020-bart} and PEGASUS \cite{Zhang2019PEGASUSPW}. Models in the \textsc{EXformer} and \textsc{Old} split were developed much earlier, such as Pointer-Generator \cite{see-etal-2017-get}. In line with \citet{tang-etal-2023-understanding}'s recommendations, our performance reporting focuses on the \textsc{FtSota} split, as it most accurately reflects the challenges of unfaithfulness in the \textit{most advanced} summarization models. \Cref{tab:aggrefact_stats} shows the statistics of \aggrefactsota~.

\vspace{-3mm}
\subsection{Baseline Models}
\label{sec:baselines}
\vspace{-1mm}

We compared \modelshort~ with the following \textit{non-LLM} methods. \textbf{\textsc{FactCC}} \cite{kryscinski-etal-2020-evaluating} is an entailment-based metric trained on synthetic data created through rule-based transformations of source document sentences. \textbf{\textsc{DAE}} \cite{goyal-durrett-2021-annotating} proposes an arc entailment approach that evaluates the factuality of each dependency arc of the generated summary independently with respect to the input article and then uses their aggregation as the overall score. Unlike previous work that reports the performance of \textsc{DAE} trained on human-annotated data, we opt for \textsc{DAE-Ent}, a variant of \textsc{DAE} trained on synthetic data, for fair comparison. \textbf{\textsc{QuestEval}} \cite{scialom-etal-2021-questeval} introduces a QA-based metric that calculates factuality by aggregating answer overlap scores from queries derived from both the input article and the summary. \textbf{\textsc{SummaC}} \cite{laban2022summac} focuses on detecting factual inconsistencies by aggregating sentence-level entailment scores between the input document and summary sentences, with (\textbf{\textsc{Conv}}) or without (\textbf{\textsc{ZS}}) further fine-tuning. \textbf{\textsc{QAFactEval}} \cite{fabbri-etal-2022-qafacteval} is a QA-based metric analogous to the precision-based component of \textsc{QuestEval} and includes optimized question answering, generation, and answer-overlap components. \textbf{\textsc{AlignScore}} \cite{Zha2023AlignScoreEF} learns a metric based on information alignment between two texts by integrating a diverse range of data sources. \textbf{\textsc{FalseSum}} \cite{utama-etal-2022-falsesum} is an entailment-based method that is trained on negative data produced by infilling masked spans of reference summaries with control code. For a fair comparison with entailment-based models, we retrain \textsc{FactCC} and \textsc{FalseSum} with \texttt{roberta-large-mnli} as the base models.

In addition, we also compared with \textit{LLM-based} methods. We tested \textbf{ChatGPT} \cite{openai2023chatgpt} using different prompts \cite{Luo2023ChatGPTAA, Wang2023IsCA}: zero-shot binary rating (\textbf{\textsc{ZS}}), zero-shot binary rating with chain-of-thought (\textbf{\textsc{CoT}}), direct assessment on a continuous scale from 0 to 100 (\textbf{\textsc{DA}}), and direct assessment on a discrete scale of one-to-five (\textbf{\textsc{Star}}). \textbf{\textsc{G-Eval}} \cite{Liu2023GEvalNE} uses chain-of-thoughts and a form-filling template to produce a discrte scale of one-to-five. We implement it with GPT-4 Turbo (\texttt{gpt-4-1106-preview}).

\vspace{-3mm}
\subsection{Evaluation Criteria}
\label{sec:eval_criteria}
\vspace{-1mm}

We use balanced accuracy \cite{Brodersen2010TheBA} as our evaluation metric. Then, we set a threshold for each model such that the threshold optimizes its performance on the validation split, following \citet{laban2022summac}. Therefore, our models will produce a binary label instead of a scalar value.

\vspace{-3mm}
\subsection{Threshold Selection for \negfilter~}
\label{sec:threshold}
\vspace{-1mm}

The thresholds for our \negfilter~ module, $\tau_1$ and $\tau_2$, are meticulously chosen to reach the best balanced accuracy on the validation split of \aggrefactsota~ dataset. The selection criteria aim to strike a delicate balance: eliminating low-quality negative samples while preserving a rich variety of error types within the training dataset. For optimal results, we configured $\tau_1$ to 0.9 and $\tau_2$ to -1.8. \looseness=-1

\section{Results and Discussion}
\vspace{-2mm}

\subsection{Main Results}
\vspace{-2mm}

\begin{table}[t]
\small
\centering
\begin{adjustbox}{max width=0.49\textwidth}
{
\begin{tabular}{lccc}
\toprule
 & \begin{tabular}[c]{@{}c@{}}\textbf{\textsc{AggreFact-}}\\\textbf{\textsc{Cnn-FtSota}}\end{tabular} & \begin{tabular}[c]{@{}c@{}}\textbf{\textsc{AggreFact-}}\\\textbf{\textsc{XSum-FtSota}}\end{tabular} & \textsc{Avg}\\
\midrule
\multicolumn{4}{c}{\textit{Non-LLM-based}} \\
\midrule
\textsc{DAE} & 65.0 $\pm$ 3.5 & 62.3 $\pm$ 1.9 & 63.7\\
\textsc{QuestEval} & 70.2 $\pm$ 3.2 & 59.5 $\pm$ 2.7 & 64.9 \\
\textsc{SummaC-ZS} & 64.0 $\pm$ 3.8 & 56.4 $\pm$ 1.2 & 60.2 \\
\textsc{SummaC-Conv} & 61.0 $\pm$ 3.9 & 65.0 $\pm$ 2.2 & 63.0 \\
\textsc{QAFactEval} & 67.8 $\pm$ 4.1 & 63.9 $\pm$ 2.4 & 65.9 \\
\textsc{FactCC} & 57.6 $\pm$ 3.9 & 57.2 $\pm$ 1.7 & 57.4 \\
\textsc{FalseSum} & 50.5 $\pm$ 3.3 & 54.7 $\pm$ 1.9 & 52.6 \\
\textsc{AlignScore} & 67.0 $\pm$ 3.1 & 60.3 $\pm$ 1.9 & 63.7\\

\midrule
\multicolumn{4}{c}{\textit{LLM-based}} \\
\midrule
\textsc{ChatGPT-ZS} & 56.3 $\pm$ 2.9 & 62.7 $\pm$ 1.7 & 59.5 \\
\textsc{ChatGPT-COT} & 52.5 $\pm$ 3.3 & 55.9 $\pm$ 2.1 & 54.2 \\
\textsc{ChatGPT-DA} & 53.7 $\pm$ 3.5 & 54.9 $\pm$ 1.9 & 54.3 \\
\textsc{ChatGPT-Star} & 56.3 $\pm$ 3.1 & 57.8 $\pm$ 0.2 & 57.1\\
\textsc{G-Eval} &  69.9 $\pm$ 3.5 & \textbf{65.8 $\pm$ 1.9} & 67.9\\
\midrule
\modelshort~ (ours) & \textbf{72.3 $\pm$ 2.5} & 64.1 $\pm$ 1.8 & \textbf{68.2} \\
\bottomrule
\end{tabular}
}
\end{adjustbox}
\vspace{-2mm}
\caption{Balanced binary accuracy (\%) on the \aggrefactsota~ test set. We show 95\% confidence intervals. Highest performance is highlighted in \textbf{bold}. The \textsc{Avg} score is computed by taking the arithmetic average of the performance on \textsc{AggreFact-Cnn-FtSota} and \textsc{AggreFact-XSum-FtSota}.
} \label{tab:sota_eval}
\vspace{-4mm}
\end{table}

The summarized results in \Cref{tab:sota_eval} indicates that our model \modelshort~ sets a new benchmark on the \aggrefactsota~ test set. We report the performance on \cnndm~ and \xsum~ using separate thresholds, following \citet{tang-etal-2023-understanding}. \modelshort~ establishes a new state-of-the-art score of 72.3\% on the \cnndm~ split, outperforming the nearest competitor by a margin of 2.1\%. On \xsum~ split, our model's performance is competitive, tying with the highest score at 64.1\%. With an overall average score of 68.2\% across both datasets, we outperform the previous best model by 2.3\% over all non-LLM approaches. The improvements underscore the advantage of incorporating AMR in generating training data and the strategic selection of high-quality negative data through our data selection module.

\begin{table}
\small
\centering
\begin{adjustbox}{max width=0.49\textwidth}
{
\begin{tabular}{lccc}
\toprule
  & \begin{tabular}[c]{@{}c@{}}\textbf{\textsc{AggreFact-}}\\\textbf{\textsc{Cnn-FtSota}}\end{tabular} & \begin{tabular}[c]{@{}c@{}}\textbf{\textsc{AggreFact-}}\\\textbf{\textsc{XSum-FtSota}}\end{tabular} & \textsc{Avg}\\
\midrule
\textsc{FactCC} & \gca{57.6} $\pm$ 3.9 & \gca{57.2} $\pm$ 1.7 & 57.4 \\
~~~ + Filtering & \gca{67.9} $\pm$ 2.3 & \gca{63.8} $\pm$ 2.2 & 65.8 \\
\midrule
\textsc{FalseSum} & \gca{50.5} $\pm$ 3.3 & \gca{54.7} $\pm$ 1.9 & 52.6 \\
~~~ + Filtering & \gca{52.3} $\pm$ 1.8 & \gca{59.7} $\pm$ 2.2 & 56.0 \\
\midrule
\modelshort~ (ours) & \textbf{\gca{72.3} $\pm$ 2.5} & \textbf{\gca{64.1} $\pm$ 1.8} & \textbf{68.2} \\
~~~ - Filtering & \gca{64.4} $\pm$ 3.0 & \gca{57.8} $\pm$ 2.0 & 61.1 \\
\bottomrule
\end{tabular}
}
\end{adjustbox}
\vspace{-2mm}
\caption{Balanced binary accuracy (\%) on the \textsc{AggreFact-Cnn-FtSota} and \textsc{AggreFact-XSum-FtSota} test set with or without our invalid negative data filtering module. } 
\vspace{-6mm}
\label{tab:selection_eval}

\end{table}

\vspace{-2mm}
\subsection{Impact of Our Filtering Module}

To validate the \textit{effectiveness} and \textit{generalizability} of our invalid negative data filtering component \negfilter~, we apply this module to the data generated by \textsc{FactCC} and \textsc{FalseSum} and re-train a \textsc{RoBERTa} on these newly filtered data. Additionally, we train another \textsc{RoBERTa} on our generated data without the filtering component. The results are summarized in \Cref{tab:selection_eval}. Specifically, we use all training samples from \textsc{FactCC} (1,003,355 samples), \textsc{FalseSum} (287,884 samples), and \textsc{AMRFact} (283,140 samples) to train models without filtering, respectively. Furthermore, the \textsc{FalseSum} and \textsc{FactCC} datasets retain 25,258 and 81,270 training samples, respectively, after undergoing the same filtering process as \textsc{AMRFact} (\textit{i.e.}, same $\tau_1$ and $\tau_2$). We observe a substantial performance gain when all three data generation methods incorporate the proposed invalid data filtering module. Specifically, \textsc{FactCC}'s balanced accuracy improves from 57.6\% to 67.9\% on the \cnndm~ and from 57.2\% to 63.8\% on \xsum~, while \textsc{FalseSum} sees an enhancement from 50.5\% to 52.3\% on \cnndm~ and from 54.7\% to 59.7\% on \xsum~. These improvements affirm the effectiveness and generalizability of our approach in filtering invalid negative training data. Conversely, removing the filtering from our \modelshort~ data causes a decrease in balanced accuracy, dropping from 72.3\% to 64.4\% on \cnndm~ and from 64.1\% to 57.8\% on \xsum~, resulting in an average drop from 68.2\% to 61.1\%. This stark contrast not only underlines the critical role of our invalid negative data filtering module in enhancing the performance of inconsistency detection models, supporting our hypothesis mentioned in \Cref{sec:negative-examples-selection} that \textbf{training inconsistency detection models on invalid negative data hurts the performance} but also emphasize our proposed method's ability to achieve better performance with \textbf{fewer training examples}.

\vspace{-1mm}
\subsection{Ablation Studies}
\vspace{-1mm}

In the next step, we focus on the impact of five specific perturbations on metric performance, using the \aggrefactsota~ dataset for an in-depth ablation analysis. This involves removing each perturbation from the \modelshort~ dataset to observe how their absence influences the creation of negative examples. As outlined in \Cref{tab:ablation_eval}, the results indicated a significant decrease in metric accuracy when any perturbation was excluded, highlighting their importance in enhancing metric precision.

The most notable finding is the substantial decline in performance upon removing discourse link error, suggesting that current models frequently struggle with this issue. By eliminating this perturbation, the model cannot access such negative examples during training, which significantly limits its ability to detect such inconsistencies during inference time. Conversely, omitting entity and out-of-article errors from the perturbations have a minimal impact on \cnndm~ and \xsum~, respectively, indicating the relative robustness of current models against these specific issues.

\begin{table}
\centering
\begin{adjustbox}{max width=0.49\textwidth}
{
\begin{tabular}{lccc}
\toprule
  & \begin{tabular}[c]{@{}c@{}}\textbf{\textsc{AggreFact-}}\\\textbf{\textsc{Cnn-FtSota}}\end{tabular} & \begin{tabular}[c]{@{}c@{}}\textbf{\textsc{AggreFact-}}\\\textbf{\textsc{XSum-FtSota}}\end{tabular} & \textsc{Avg}\\
\midrule
\textsc{AMRFact}            & \textbf{72.3 $\pm$ 2.5} & 64.1 $\pm$ 1.8 & \textbf{68.2} \\
~~~ - Predicate Error       & 70.0 $\pm$ 3.1 & 62.0 $\pm$ 1.9 & 66.0 \\
~~~ - Entity Error          & 70.8 $\pm$ 3.1 & 62.1 $\pm$ 2.1 & 66.5 \\
~~~ - Circumstance Error    & 68.2 $\pm$ 2.9 & 61.6 $\pm$ 1.9 & 64.9 \\
~~~ - Discourse Link Error  & 65.0 $\pm$ 2.7 & 61.7 $\pm$ 1.9 & 63.4 \\
~~~ - Out of Article Error  & 66.8 $\pm$ 2.4 & 62.3 $\pm$ 2.0 & 64.6 \\
\bottomrule
\end{tabular}
}
\end{adjustbox}
\vspace{-2mm}
\caption{Balanced binary accuracy (\%) on the \textsc{AggreFact-Cnn-FtSota} and \textsc{AggreFact-XSum-FtSota} test set \textit{without} a specific error type.} \label{tab:ablation_eval}
\vspace{-6mm}
\end{table}

\begin{table*}[h]
\small
\centering
\begin{tabular}{p{0.95\linewidth}}
\toprule

\textbf{Article}: Lord Janner signed a letter saying he wanted to remain a peer just a week \hlc{lightblue}{before} he was ruled unfit to face child sex charges. Abuse campaigners last night angrily questioned why the suspected paedophile was able to remain in the House of Lords if he was too frail to be brought before court...\\
 \midrule
\textbf{Summary}: Lord Janner signed letter saying he wanted to remain a peer on April 9. Comes a week \hlc{lightred}{after} he was ruled unfit to face child sex charges...\\

\bottomrule
\end{tabular}
\vspace{-3mm}
\caption{An example with a discourse link error from \aggrefactsota~ where models trained on \modelshort~ successfully classify it as factual inconsistent with the input article but fails when trained on \textsc{FalseSum}.}
\vspace{-4mm}
\label{tab:qualitative_analysis}
\end{table*}

\subsection{Qualitative Analysis}
\label{subsec:qualitative_analysis}

The following qualitative analysis provides insights into our model's capability to produce coherent negative examples while ensuring extensive coverage of various error types.

\begin{figure}[t]
 \centering
 \includegraphics[width=0.9\linewidth]{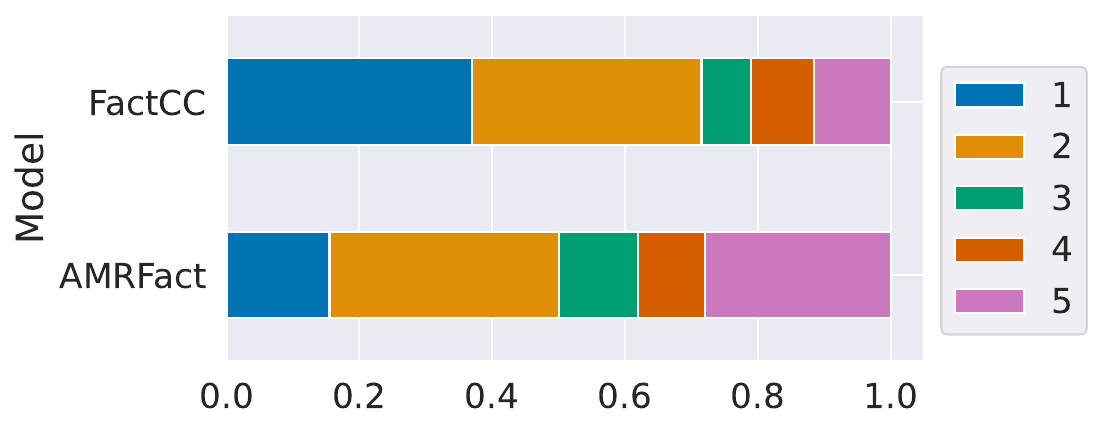}
 \vspace{-2mm}
 \caption{A breakdown of coherence scores for negative summaries produced by \modelshort~ and \textsc{FactCC}.}
 \vspace{-5mm}
 \label{fig:coherence_scores}
\end{figure}

\paragraph{Coherence}
To demonstrate that our approach produces more coherent negative summaries than entity-based baselines, we compare 200 summaries generated by \modelshort~ and \textsc{FactCC}. We use GPT-4 Turbo to assess the coherence of each summary on a scale of 1-5, where 1 indicates the least coherent and 5 means the most. The average coherence scores for \modelshort~ and \textsc{FactCC} are 3.01 and 2.24, respectively, highlighting the advantages of our approach in generating more coherent negative summaries compared to \textsc{FactCC}. We show the breakdown score in \Cref{fig:coherence_scores}.

\paragraph{Error-type coverage}

To verify that generation-based baselines suffer from the insufficient error-type coverage issue, we evaluate the error type distribution of the negative data produced by \textsc{FalseSum} by sampling its generated summaries and query GPT-4 Turbo to determine the error type within each summary. The findings revealed a notable gap: only 5 out of 1,000 summaries exhibited a Discourse Link Error. This scarcity of training data for this error type could lead to models trained on such data failing to detect these errors reliably. In \Cref{tab:qualitative_analysis}, we show an example with a Discourse Link Error in which our model predicts the correct label and \textsc{FalseSum} fails, underscoring the issue of inadequate error-type coverage. 
The prompts used for both analyses are detailed in \Cref{apx:prompt_details}.

\subsection{Remaining Challenges}
\label{sec:remaining_challenges}

\begin{figure}[b]
 \centering
\vspace{-2mm}\includegraphics[width=0.8\linewidth]{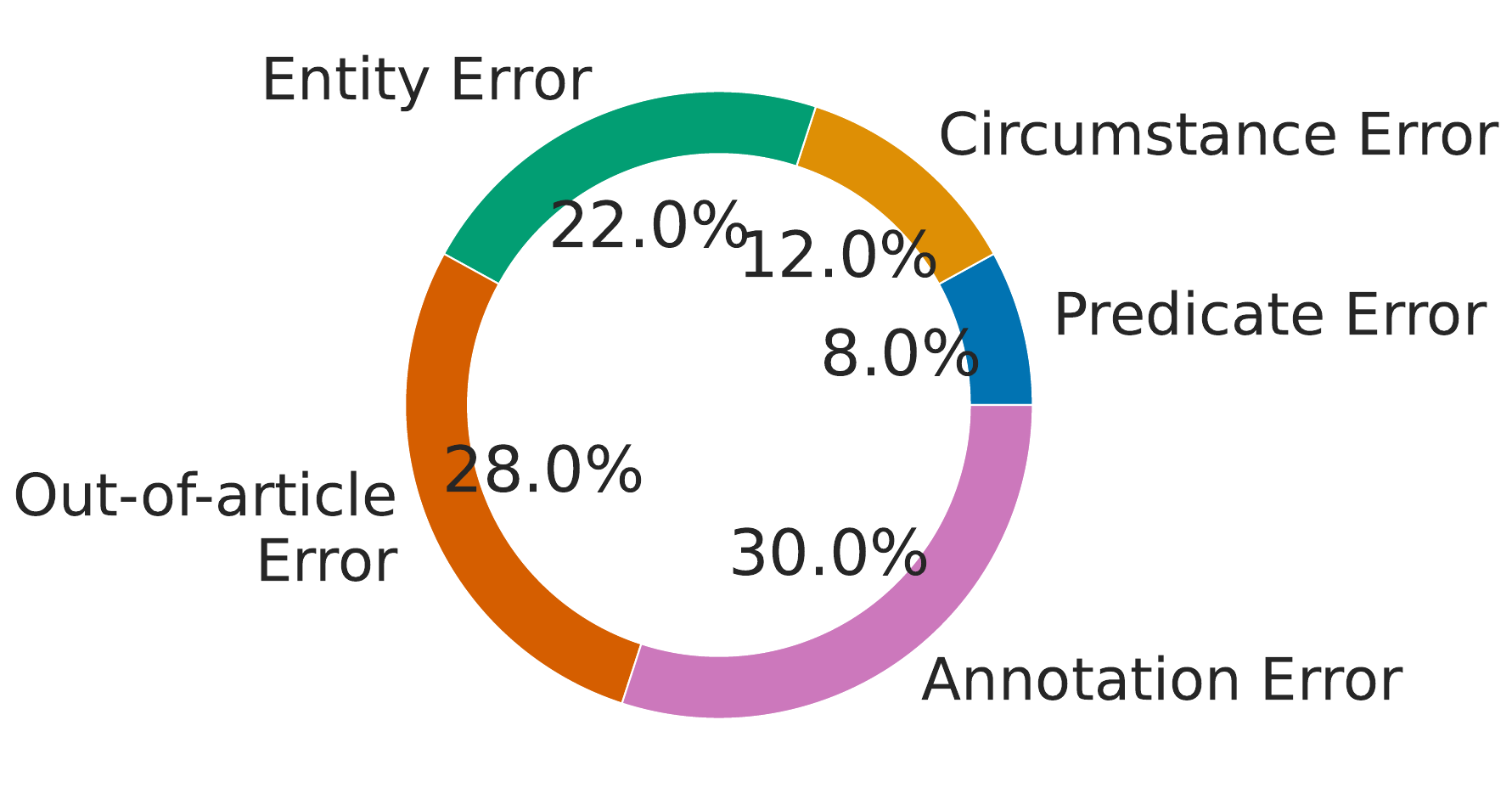}
 \vspace{-3mm}
 \caption{Distribution of the errors that our model fails to identify. The most dominant type of error is the annotation error. And errors stemming from out-of-article content were another significant source of inaccuracies.}
 \vspace{-3mm}
 \label{fig:remaining_challenges}
\end{figure}

To gain deeper insights into our model's limitations and remaining challenges, we conducted a detailed manual analysis. This involved examining a random sample of 50 examples from \aggrefactsota~ where our model incorrectly labeled factually inconsistent instances. The results of this analysis are detailed in \Cref{fig:remaining_challenges}. From this thorough assessment, several key findings emerged. First, the most dominant type of error (30$\%$) is the annotation error. This underscores the challenges in creating completely reliable benchmarks through crowd-sourced methods, a complexity also noted in the work of \citet{laban2023llms}. Second, errors stemming from out-of-article content were another significant source of inaccuracies. Intriguingly, upon further investigation, we discovered that a notable portion (8 out of 14) of these out-of-article errors were, in fact, factually accurate \cite{dong-etal-2022-faithful}. For instance,

\begin{quote}
    \small
    Former Wales captain Martyn Williams says Dan Biggar's decision to sign a new four-year contract with Ospreys will benefit the region.
\end{quote}

The above summary was labeled as unfaithful because the corresponding source document (see \Cref{tab:factual_unfaithful_article} in Appendix) does not mention that Martyn Williams was a captain. Nonetheless, this information is factually correct, as evidenced by a Wikipedia search. This discrepancy suggests that our model, in this case, might have utilized its parametric knowledge learned during the pre-training phase of \textsc{RoBERTa} rather than the content of the article itself. To mitigate this issue, future research efforts can look into creating negative training examples that consist of out-of-article errors which, although truthful according to worldly facts, are unfaithful to the source document.
\vspace{-2mm}
\section{Related Work}
\vspace{-1mm}

\subsection{Factual Consistency Metrics}
\vspace{-1mm}

Research on factual consistency metrics can be classified into QA-based and entailment-based approaches. QA-based metrics generally leverage question generation and answering components to evaluate factual consistency through the comparison of information units taken from summaries and their original sources \cite{wang-etal-2020-asking, scialom-etal-2021-questeval, fabbri-etal-2022-qafacteval, factscore}. Entailment-based methods predict whether a summary is entailed by its source article with document-sentence entailment models trained on either synthetic \cite{kryscinski-etal-2020-evaluating, goyal-durrett-2020-evaluating, yin-etal-2021-docnli, utama-etal-2022-falsesum, soleimani-etal-2023-nonfacts} or human-annotated data \cite{goyal-durrett-2021-annotating,ribeiro-etal-2022-factgraph, chan-etal-2023-interpretable}. Besides, \citet{Gekhman2023TrueTeacherLF} propose a method for generating synthetic data by annotating diverse model-generated summaries using an LLM. Alternatively, \citet{laban2022summac} break documents and summaries into sentences and employ traditional sentence-level NLI models. Moreover, \citet{feng2023factkb} utilize facts extracted from external knowledge bases to improve the generalization across domains. Our approach employs an entailment-based method, capitalizing on AMR-based perturbations to generate negative examples, which ensures high error-type coverage and synthetic data quality. Furthermore, we enhance the data quality with \negfilter~ by filtering out invalid negative samples.

\vspace{-2mm}
\subsection{AMR-based Evaluation}
\vspace{-1mm}

Few studies have attempted to employ AMR for evaluation. \citet{ribeiro-etal-2022-factgraph} parse both the input documents and summaries into AMR graphs and train a classifier to identify erroneous edges in the summary graph. However, their approach relies on human-annotated data for effective training signals. On the contrary, we parse summaries into AMR graphs and create negative training data by applying perturbations to the graphs and converting the perturbed graphs back to summaries. \citet{ghazarian-etal-2022-deam} uses a similar approach to generate training data for a dialogue coherence evaluation metric. Notably, their perturbation techniques focused on coherence evaluation, and thus ill-suited for factual consistency evaluation. By contrast, we design our perturbation method by taking inspiration from common factual errors in summaries produced by state-of-the-art summarization systems. \looseness=-1
\vspace{-1mm}
\section{Conclusion}
\vspace{-2mm}

We introduce a novel framework, \modelshort~, which leverages Abstract Meaning Representations (AMRs) to generate factually inconsistent summaries for summarization factuality evaluation. Our method addresses the prevalent issues of low coherence and insufficient error-type coverage observed in prior entailment-based approaches. By parsing factually consistent reference summaries into AMR graphs and injecting controlled factual inconsistencies, we successfully create coherent but factually inconsistent summaries with broad error-type coverage. The introduction of the \negfilter~ module further enhances the quality of the generated negative samples. The effectiveness and generalizability of this approach are validated through its state-of-the-art performance on the \aggrefactsota~ dataset. Our contributions not only advance the field in generating high-quality, factually inconsistent summaries but also provide a scalable and efficient solution for enhancing the factuality evaluation of summarization systems. The results underscore the potential of AMR-based perturbations in improving the integrity and reliability of natural language generation.
\section{Ethical Considerations}

The metrics introduced in this study are based on models trained predominantly with English language documents. Consequently, they reflect the cultural nuances and perspectives common in English-speaking societies. It is important to acknowledge the potential presence of political and gender biases within these datasets. The models, and by extension, the metrics derived from them, might inadvertently perpetuate these biases \cite{sun-etal-2022-bertscore,qiu-etal-2023-gender}. We have not rigorously evaluated these metrics for bias-related issues. Therefore, we advise users to be mindful of these limitations when applying these metrics, considering the potential for underlying biases in their use and interpretation. On a positive note, our methodology can serve as a valuable tool for detecting factually inconsistent summaries and making the summaries more factual.

\section{Limitations}

When employed effectively, the metrics outlined in this paper can serve as valuable tools for identifying errors in summarization models. However, it is crucial to recognize that these metrics are not infallible in detecting every factual inconsistency. This limitation should be considered when using these metrics to assess summaries for downstream applications. Moreover, our goal is to demonstrate the potential of using AMR-based perturbations for generating coherent yet intentionally factually inconsistent summaries. However, we acknowledge that the quality of our perturbations depends on pre-trained text-to-AMR parsers and AMR-to-Text generators. If these models are not strong, our summaries may suffer in quality, as discussed in \Cref{sec:negative-examples-selection}. Therefore, it is essential to be aware of these constraints because factual inconsistencies in summaries can potentially propagate misinformation online. Therefore, while these metrics are helpful, their limitations in fully capturing factual inconsistencies must be acknowledged and managed carefully.
\section{Acknowledgment}

We thank anonymous reviewers for their helpful feedback. We also thank Zi-Yi Dou, Po-Nien Kung, Chujie Zheng, Haw-Shiuan Chang, Di Wu, and other members from the UCLA NLP group for their feedback and discussions. This research is supported by Meta Sponsor Research Award, Okawa Foundation Research Grant, Amazon Alexa AI Research Award, and a gift from UCLA Institute for Technology, Law and Policy.

\bibliography{anthology,custom}
\bibliographystyle{acl_natbib}

\clearpage
\appendix

\section{Experimental Settings}

We fine-tune \texttt{roberta-large-mnli} on our \modelshort~ dataset for 10 epochs with a learning rate of 5E-05.

\section{Main Results Breakdown}

\Cref{tab:results_breakdown} demonstrates \modelshort~’s performance on the diverse member datasets within \aggrefactsota~. These experiments demonstrate that \modelshort~ outperforms \textsc{SummaC-ZS} and \textsc{QAFactEval} on five out of seven datasets within \aggrefactsota~.

\begin{table*}[b]
\small
\centering
\begin{tabular}{lcccccccc}
\toprule
 & Polytope & SummEval & FRANK & Wang'20 & \multicolumn{2}{c}{CLIFF} & Goyal'21 & Cao'22\\
\midrule

\textsc{Split} & \textsc{Cnn} & \textsc{Cnn} & \textsc{Cnn} & \textsc{Xsum} & \textsc{Cnn} & \textsc{Xsum} & \textsc{Xsum} & \textsc{Xsum} \\

\textsc{\# of sample} & 34 & 200 & 175 & 119 & 150 & 150 & 50 & 239 \\
\midrule

\modelshort~ & \textbf{1.0} & \textbf{0.807} & \textbf{0.724} & 0.595 & 0.710 & \textbf{0.667} & 0.591 & \textbf{0.645} \\

\textsc{SummaC-ZS} & 0.971 & 0.622 & 0.570 & 0.698 & 0.656 & 0.596 & 0.466 & 0.490 \\

\textsc{QAFactEval} & 0.324 & 0.652 & 0.547 & \textbf{0.756} & \textbf{0.716} & 0.626 & 0.754 & 0.613 \\

\bottomrule
\end{tabular}
\caption{Performance breakdown on the diverse member datasets within \aggrefactsota~.} 
\label{tab:results_breakdown}
\end{table*}

\begin{table*}[b]
\small
\centering
\begin{tabular}{p{0.9\linewidth}}
\toprule
The 26-year-old fly-half has agreed a new deal which could keep him in Wales until the 2019 World Cup. Williams, who played 100 times for Wales, believes it will help Ospreys keep and recruit players. "It's fabulous news for Welsh rugby and the Ospreys in particular," he said on BBC Wales' Scrum V TV programme. "Not only is Dan committing himself to Wales for the next four years, but it helps the Ospreys with recruitment for the next couple of seasons. "If players were looking to sign and come to the Ospreys if they can see Dan Biggar is going to be there for the next three or four seasons that helps them as well." Biggar's current deal was due to expire at the end of this season. He is the first of Wales' 17 dually contracted players to re-sign on the deals which are 60\% funded by the WRU and 40\% by the region. In addition to potentially attracting new players to the region, Biggar's decision to stay may help negotiations with his Ospreys and Wales colleagues scrum-half Rhys Webb and second row Alun Wyn Jones. Both players' contracts expire in the summer of 2016, with Ospreys skipper Jones saying in November that he was still weighing up his options. Scarlets are in talks with Wales centre Scott Williams over extending his dual contract, and have secured the return of British and Irish Lions centre Jonathan Davies from Clermont Auvergne next season.\\

\bottomrule
\end{tabular}
\caption{The source document corresponding to the factual but unfaithful summary mentioned in \Cref{sec:remaining_challenges}.}
\label{tab:factual_unfaithful_article}
\end{table*}

\section{Prompt Details}
\label{apx:prompt_details}

We show the prompt used to analyze the error distribution for \textsc{FalseSum} in \Cref{tab:falsesum_error_dist_prompt} and coherence evaluation in \Cref{tab:coherence_eval}.

\begin{table*}[h]
\small
\centering
\begin{tabular}{p{0.95\linewidth}}
\toprule
You will be given one reference summary and one generated summary written for a news article.\\

Your task is to determine the type of factual error in the generated summary with regard to the article.\\
\\
Please make sure you read and understand these instructions carefully. Please keep this document open while reviewing, and refer to it as needed.\\
\\
Types of Errors:\\
\\
1. Predicate Error: The predicate of the summary does not align with the information provided in the original document.\\
2. Entity Error: The primary arguments, or their attributes, associated with the predicate are incorrect.\\
3. Circumstance Error: The supplementary details, such as location or time, that define the context of a predicate are incorrect.\\
4. Discourse Link Error: Errors arise from the improper connection of statements within the discourse, such as errors in temporal ordering or causal link.\\
5. Out of Article Error: The statement conveys information that is absent in the original document.\\

\\
Evaluation Steps:\\
\\
1. Read the news article carefully and identify the main topic and key points.\\
2. Read the generated summary and compare it to the news article. \\
3. Determine the type of errors based on our definition.\\
\\

Example:\\

\\
Source Text:\\

[Document]\\

\\
Reference Summary:\\

[Reference Summary]\\

\\
Generated Summary:\\

[Generated Summary] \\

\\
Evaluation Form (error type ONLY):\\
\\
- Error Type:\\

\bottomrule
\end{tabular}
\caption{The prompt to GPT-4 Turbo for determining the error type of a non-factual summary.}
\label{tab:falsesum_error_dist_prompt}
\end{table*}
\begin{table*}[h]
\small
\centering
\begin{tabular}{p{0.95\linewidth}}
\toprule
You will be given one summary written for a news article.\\

Your task is to rate the summary on one metric.\\

Please make sure you read and understand these instructions carefully. Please keep this document open while reviewing, and refer to it as needed.\\

\\
Evaluation Criteria:\\

Coherence (1-5). Here, we focus on "within sentence coherence". It involves ensuring that the components of a single sentence – such as subjects, verbs, objects, and other elements – are logically and grammatically connected, making the sentence clear and understandable.\\

Evaluation Steps:\\

\\
1. Read the summary carefully.\\
2. Evaluate the summary based on the evaluation criteria.\\
3. Assign a score for coherence on a scale of 1 to 5, where 1 is the lowest and 5 is the highest based on the Evaluation Criteria.\\

\\
Generated Summary:\\

[Generated Summary]\\

\\
Evaluation Form (scores ONLY):\\

- Coherence:\\

\bottomrule
\end{tabular}
\caption{The prompt to GPT-4 Turbo for determining the coherence of a generated summary.}
\label{tab:coherence_eval}
\end{table*}
\end{document}